\documentclass[runningheads]{llncs}

\newcommand{\final}{1} 
\newcommand{\forReview}{0} 

\usepackage{graphicx}
\usepackage{comment}
\usepackage{amsmath,amssymb} 
\usepackage{color}
\usepackage{floatrow}
\newfloatcommand{capbtabbox}{table}[][\FBwidth]
\usepackage[width=122mm,left=12mm,paperwidth=146mm,height=193mm,top=12mm,paperheight=217mm]{geometry}

\usepackage[symbol]{footmisc}

\usepackage{epsfig}
\usepackage{graphicx}

\usepackage{amsthm}
\usepackage{subcaption}
\usepackage{color}
\usepackage{ifthen}
\usepackage{float}
\usepackage{alltt}
\usepackage{newlfont} 

\usepackage{wrapfig}
\usepackage{booktabs}
\usepackage{multirow}
\usepackage{comment}

\usepackage{comment}

\definecolor{DeltaColor}{rgb}{0.039,0.73,0.71}
\definecolor{SetaColor}{rgb}{0.867, 0.0235, 0.376}
\definecolor{SigmaColor}{rgb}{0.98,0.45,0.0}
\definecolor{RedColor}{rgb}{0.8,0,0}
\definecolor{AlphaColor}{rgb}{0,0,0.8}
\definecolor{BetaColor}{rgb}{0.8,0,0.8}
\definecolor{GammaColor}{rgb}{0.5,0,0.7}
\definecolor{EpsilonColor}{rgb}{0.353,0.725,0.906}
\definecolor{TauColor}{rgb}{0.423,0.235,0.192}
\newcommand{\weikai}[1]{{\color{RedColor} Weikai: #1 $\qed$}}
\newcommand{\heming}[1]{{\color{AlphaColor} Heming: #1 $\qed$}}
\newcommand{\xiaoguang}[1]{{\color{SigmaColor} Xiaoguang: #1 $\qed$}}
\newcommand{\yu}[1]{{\color{BetaColor} Yu: #1 $\qed$}}
\newcommand{\hang}[1]{{\color{EpsilonColor} Hnag: #1 $\qed$}}

\newcommand{\nothing}[1]{}

\definecolor{AudioColor}{rgb}{0.56,0.34,0.62}

\definecolor{DeadlineColor}{rgb}{0.9,0.4,0} 

\definecolor{figred}{rgb}{1,0,0}
\definecolor{figgreen}{rgb}{0,0.6,0}
\definecolor{figblue}{rgb}{0,0,1}
\definecolor{figpink}{rgb}{1,0.63,0.63}

\ifthenelse{\equal{\final}{1}}
{
\renewcommand{\heming}[1]{}
\renewcommand{\yu}[1]{}
\renewcommand{\hang}[1]{}
\renewcommand{\xiaoguang}[1]{}
\renewcommand{\weikai}[1]{}
\renewcommand{\note}[1]{}
}
{}

\newcounter{pccount}
\floatstyle{plain}

\newcommand{\filename}[1]{\url{#1}}
\newcommand{\foldername}[1]{\url{#1}}

\newcommand{\datasetName}{Deep Fashion3D}

\ifthenelse{\equal{\forReview}{1}}
{
}
{
}

\graphicspath{
	{figs/raster/}
	{figs/handdrawn/}
}

\begin{document}
	
	
	\pagestyle{headings}
	\mainmatter
	\def\ECCVSubNumber{2149}  
	
	\title{\datasetName{}: A Dataset and Benchmark for 3D Garment Reconstruction from Single Images}
	
	\titlerunning{Deep Fashion3D} 
	\author{
	Heming Zhu\inst{1,2,3\dag} \and
    Yu Cao\inst{1,2,4\dag} \and
    Hang Jin\inst{1,2\dag} \and 
    Weikai Chen\inst{5} \and 
    Dong Du\inst{6} \and 
    Zhangye Wang\inst{3} \and 
    Shuguang Cui\inst{1,2} \and 
    Xiaoguang Han\inst{1,2 \ddag}
    }
	\institute{
	The Chinese University of Hong Kong, Shenzhen \and  
	Shenzhen Research Institute of Big Data \and  
    State Key Lab of CAD$\&$CG, Zhejiang University\and  
	Xidian University \and   
	Tencent America \and 
	University of Science and Technology of China 
	}

	\authorrunning{H. Zhu et al.}
	
	\maketitle
	
	
\begin{abstract}
High-fidelity clothing reconstruction is the key to achieving photorealism in a wide range of applications including human digitization, virtual try-on, etc. Recent advances in learning-based approaches have accomplished unprecedented accuracy in recovering unclothed human shape and pose from single images, thanks to the availability of powerful statistical models, e.g. SMPL~\cite{loper2015smpl}, learned from a large number of body scans. In contrast, modeling and recovering clothed human and 3D garments remains notoriously difficult, mostly due to the lack of large-scale clothing models available for the research community. We propose to fill this gap by introducing \datasetName{}, the largest collection to date of 3D garment models, with the goal of establishing a novel benchmark and dataset for the evaluation of image-based garment reconstruction systems. \datasetName{} contains 2078 models reconstructed from real garments, which covers 10 different categories and 563 garment instances. It provides rich annotations including 3D feature lines, 3D body pose and the corresponded multi-view real images. In addition, each garment is randomly posed to enhance the variety of real clothing deformations. To demonstrate the advantage of \datasetName{}, we propose a novel baseline approach for single-view garment reconstruction, which leverages the merits of both mesh and implicit representations. A novel adaptable template is proposed to enable the learning of all types of clothing in a single network. Extensive experiments have been conducted on the proposed dataset to verify its significance and usefulness. We will make \datasetName{} publicly available upon publication.      
\end{abstract}


	\footnote[0]{\dag First three authors should be considered as joint first authors.}
	\footnote[0]{\ddag Xiaoguang Han is the corresponding author.  Email:hanxiaoguang@cuhk.edu.cn.}
	\section{Introduction}
\label{sec:intro}


Human digitization is essential to a variety of applications ranging from visual effects, video gaming, to telepresence in VR/AR. 
The advent of deep learning techniques has achieved impressive progress in recovering unclothed human shape and pose simply from multiple~\cite{huang2018deep,xu20193d}  or even single~\cite{natsume2019siclope,varol2018bodynet,alldieck2019tex2shape} images.
However, these leaps in performance come only when a large amount of labeled training data is available. 
Such limitation has led to inferior performance of reconstructing clothing -- the key element of casting a photorealistic digital human, compared to that of naked human body reconstruction.
One primary reason is the scarcity of 3D garment datasets in contrast with large collections of naked body scans, e.g. SMPL~\cite{loper2015smpl}, SCAPE~\cite{anguelov2005scape}, etc. 
In addition, the complex surface deformation and large diversity of clothing topologies have introduced additional challenges in modeling realistic 3D garments. 


\begin{figure}[ht]
\centering
\includegraphics[width=1.0\linewidth]{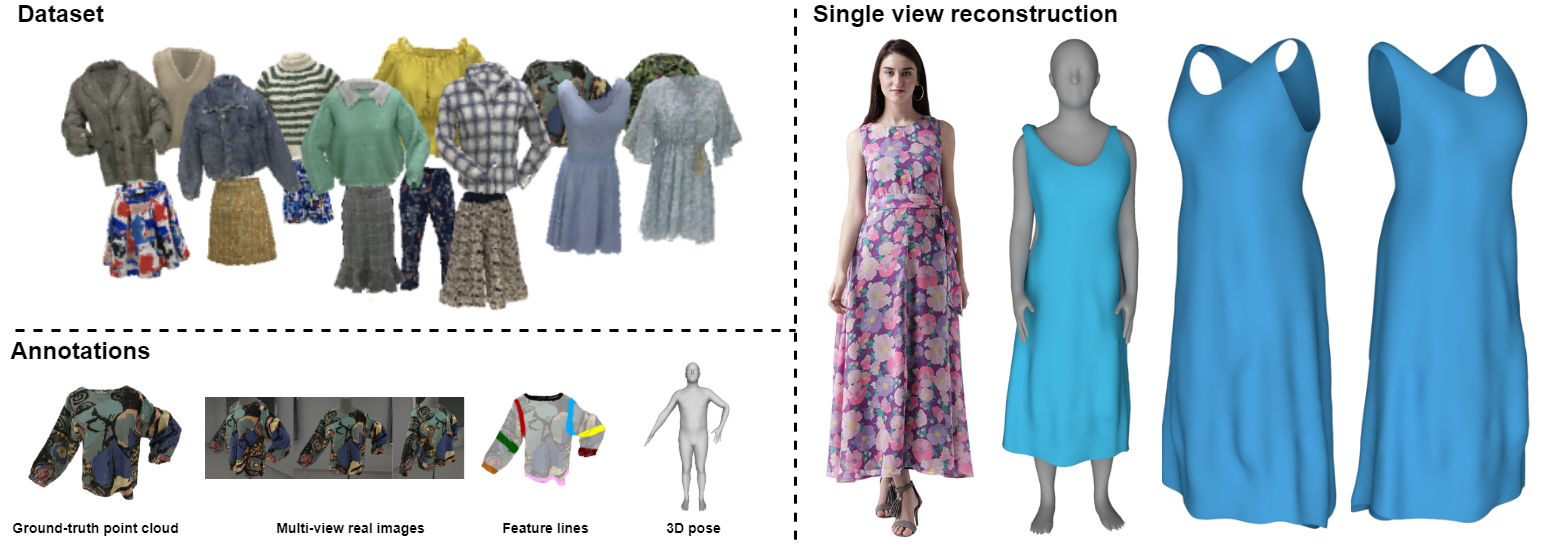}
\caption{
We present \datasetName{}, a large-scale repository of 3D clothing models reconstructed from real garments. It contains over 2000 3D garment models, spanning 10 different cloth categories. Each model is richly labeld with ground-truth point cloud, multi-view real images, 3D body pose and a novel annotation named feature lines. With \datasetName{}, inferring the garment geometry from a single image becomes possible. 
}
\label{fig:teaser}
\end{figure}


To address the above issues, there is an increasing need of constructing a high-quality 3D garment database that satisfies the following properties. First of all, it should contain a large-scale repository of 3D garment models that cover a wide range of clothing styles and topologies. 
Second, it is preferable to have models reconstructed from the real images with physically-correct clothing wrinkles to accommodate the requirement of modeling complicated dynamics and deformations caused by the body motions. Lastly, the dataset should be labeled with sufficient annotations, e.g. the corresponded real images, body pose, etc., to provide strong supervision for deep generative models.   

Multi-Garment Net (MGN)~\cite{bhatnagar2019multi} introduces the first dataset specialized for digital clothing obtained from real scans. The proposed digital wardrobe contains 356 digital scans of clothed people which are fitted to pre-defined parametric cloth templates. However, the digital wardrobe only captures 5 garment categories, which is quite limited compared to the large variety of garment styles. Apart from 3D scans, some recent works~
\cite{garmentdesign_Wang_SA18,gundogdu2019garnet} propose to leverage synthetic data
obtained from physical simulation. However, the synthetic models lack realism compared to the 3D scans and cannot provide the corresponding real images, which are critical to generalizing the trained model to images in the wild.


In this paper, we address the lack of data by introducing \datasetName{}, the largest 3D garment dataset by far, that contains thousands of 3D clothing models with comprehensive annotations. 
Compared to MGN, the collection of \datasetName{} is one order of magnitude larger -- including 2078 3D models reconstructed from real garments. 
It is built from 563 diverse garment instances, covering 10 different clothing categories. 
Annotation-wise, we introduce a new type of annotation tailored for 3D garment -- 3D feature lines.
The feature lines denote the most prominent geometrical features on garment surfaces (see Fig.~\ref{fig:featureLine}), including necklines, cuff contours, hemlines, etc, which provide strong priors for 3D garment reconstruction.
Apart from feature lines, our annotations also include calibrated multi-view real images and the corresponded 3D body pose. 
Furthermore, each garment item is randomly posed to enhance the dataset capacity of modeling dynamic wrinkles.

To fully exploit the power of \datasetName{}, we propose a novel baseline approach that is capable of inferring realistic 3D garments from a single image. 
Despite the large diversity of clothing styles, most of the existing works are limited to one fixed topology~\cite{danvevrek2017deepgarment,jin2018pixel}.
MGN~\cite{bhatnagar2019multi} introduces class-specific garment network -- each deals with a particular topology and is trained by one-category subset of the database. 
However, given the very limited data, each branch is prone to having overfitting problems. 
We propose a novel representation, named adaptable template, that can scale to varying topologies during training. It enables our network to be trained using the entire dataset, leading to stronger expressiveness.
Another challenge of reconstructing 3D garments is that clothing model is typically a shell structure with open boundaries. Such topology can hardly be handled by the implicit or voxel representation. 
Yet, the methods based on deep implicit functions~\cite{mescheder2019occupancy,park2019deepsdf} have shown their ability of modeling fine-scale deformations that the mesh representation is not capable of.
We propose to connect the good ends of both worlds by transferring the high-fidelity local details learnt from implicit reconstruction to the template mesh with correct topology and robust global deformations. 
In addition, since our adaptable template is built upon the SMPL topology, it is convenient to repose or animate the reconstructed results.
The proposed framework is implemented in a multi-stage manner with a novel feature line loss to regularize mesh generation.

We have conducted extensive benchmarking and ablation analysis on the proposed dataset. Experimental results demonstrate that the proposed baseline model trained on \datasetName{} sets new state of the art on the task of single-view garment reconstruction.
Our contributions can be summarized as follows:

\begin{itemize}
    \item We build \datasetName{}, a large-scale, richly annotated 3D clothing dataset reconstructed from real garments. To the best of our knowledge, this is the largest dataset of its kind. 
  
    \item We introduce a novel baseline approach that combines the merits of mesh and implicit representation and is able to faithfully reconstruct 3D garment 
    from a single image.
   
    \item We propose a novel representation, called adaptable template, that enables encoding clothing of various topologies in a single mesh template. 
     
    \item We first present the feature line annotation specialized for 3D garments, which can provide strong priors for garment reasoning related tasks, e.g., 3D garment reconstruction, classification, retrieval, etc.  
    
    \item We build a benchmark for single-image garment reconstruction by conducting extensive experiments on evaluating a number of state-of-the-art single-view reconstruction approaches on \datasetName{}. 
\end{itemize}





	\section{Related Work}
\label{sec:related_work}

\paragraph{3D Garment Datasets.} While most of existing repositories focus on naked~\cite{anguelov2005scape,Bogo:CVPR:2014,loper2015smpl,dfaust:CVPR:2017} or clothed~\cite{Zheng2019DeepHuman} human body, datasets specially tailored for 3D garment is very limited.
BUFF dataset~\cite{zhang2017detailed} consists of high-resolution 4D scans of clothed human with very limited ammount. In addition, it fails to provide separated models for body and clothing.
Segmenting garment models from the 3D scans remains extremely laborious and often leads to corrupted surfaces due to occlusions. To address this issue, Pons-Moll et al.~\cite{ponsmollSIGGRAPH17clothcap} propose an automatic solution to extract the garments and their motion from 4D scans.
Recently, a few datasets specialized for 3D garment are proposed. Most of the works~\cite{gundogdu2018garnet,garmentdesign_Wang_SA18} propose to synthetically generate garment dataset using physical simulation. However, the quality of the synthetic data is not on par with that of real data. In addition, it remains difficult to generalize the trained model to real images as only synthetic images are available.
MGN~\cite{bhatnagar2019multi} introduces the first garment dataset obtained from 3D scans. However, the dataset only covers 5 cloth categories and is limited to a few hundreds of samples. In contrast, \datasetName{} collects more than two thousand clothing models reconstructed from real garments, which covers a much larger diversity of garment styles and topologies. Further, the novel annotation of feature lines provides stronger and more accurate supervision for reconstruction algorithms, which is demonstrated in Section~\ref{sec:result}.

\paragraph{Performance capture.} 
Over the past decades, progress \cite{wang2011data,miguel2012data,matsuyama20123d} has been made to capture cloth surface deformation in motion. 
Vision-based approaches strive to leverage the easily accessible RGB data and develop frameworks either based on texture pattern tracking~\cite{white2007capturing,scholz2005garment}, shading cues~\cite{zhou2013garment} or calibrated silhouettes obtained from multi-view videos~\cite{carranza2003free,starck2007surface,leroy2017multi,cagniart2010probabilistic}. However, without dense correspondences or priors, the silhouette-based approaches cannot fully recover the fine details. To improve the reconstruction quality, stronger prior knowledge, including the clothing type~\cite{de2008performance}, pre-scanned template model~\cite{habermann2019livecap}, stereo~\cite{bradley2008markerless} and photometric~\cite{hernandez2007non,vlasic2009dynamic} constraints, has been considered in recent works. 
With the advances of fusion-based solutions~\cite{izadi2011kinectfusion,newcombe2015dynamicfusion}, template model can be eliminated as the surface geometry can be progressively fused on the fly~\cite{collet2015high,dou2016fusion4d} with even a single depth camera~\cite{yu2019simulcap,yu2018doublefusion,yu2017bodyfusion}.
Yet, most of the existing works estimate body and clothing jointly and thus cannot obtain a separated cloth surface from the output.
Chen et al.~\cite{chen2015garment} propose to model 3D garment from a single depth camera by fitting deformable templates to the initial mesh generated by KinectFusion.
 

\paragraph{Single-view garment reconstruction.}
Inferring 3D cloth from a single image is highly challenging due to the scarcity of the input and the enormous searching space. Statistical model has been introduced for such ill-posed problem to provide strong priors for plausible prediciton. However, most models ~\cite{anguelov2005scape,loper2015smpl,hasler2009statistical,pons2015dyna,joo2018total} are restricted to capturing human body only. 
Attempts have been made to jointly reconstruct body and clothing from videos~\cite{alldieck20183DV,alldieck2018video} and multi-view images~\cite{huang2018deep,xu20193d}.
Recent advances in deep learning based approaches~\cite{natsume2019siclope,varol2018bodynet,saito2019pifu,alldieck2019tex2shape,lazova3dv2019,alldieck19cvpr,pumarola20193dpeople,chen2013deformable,tang2019neural} have achieved single-view clothed body reconstruction.
However, for all these methods, tedious manual post-processing is required to extract the clothing surface from the reconstructed result. And yet, the reconstructed clothing lacks realism.
DeepWrinkles~\cite{lahner2018deepwrinkles} proposes an approach to synthesizes faithful clothing wrinkles onto a coarse garment mesh following a given pose.
Jin et al.~\cite{jin2018pixel} leverage similar idea with \cite{Huynh_2018_CVPR}, which encodes detailed geometry deformations in the uv space. However, the method is limited to a fixed topology and cannot scale well to large deformations.
Dan{\v e}{\v r}ek et al.~\cite{danvevrek2017deepgarment} propose to use physics based simulations as supervision for training a garment shape estimation network. However, the quality of their results is limited to that of the synthetic data and thus cannot achieve high photo-realism.
Closer to our work, Multi-Garment Net~\cite{bhatnagar2019multi} learns per-category garment reconstruction from images using 3D scanned data. Nonetheless, their method typically requires 8 frames as input while our approach only consumes a single image. Further, since MGN relies on pre-trained parametric models, it cannot deal with out-of-scope deformations, especially the clothing wrinkles that are dependent on body poses. In contrast, our approach is blendshape-free and is able to faithfully capture multi-scale shape deformations.


	\section{Dataset Construction}
\label{sec:dataset}

Despite the rapid evolution of 2D garment image datasets from DeepFashion~\cite{liuLQWTcvpr16DeepFashion} to DeepFashion2~\cite{ge2019deepfashion2} and FashionAI~\cite{zou2019fashionai}, large-scale collection of 3D clothing is very rare. 
The digital wardrobe released by MGN~\cite{bhatnagar2019multi} only contains 356 scans and is limited to only 5 garment categories, which is not sufficient for training an expressive reconstruction model. 
To fill this gap, we build a more comprehensive dataset named \datasetName{}, which is one order larger than MGN, richly annotated and covers a much larger variations of garment styles.
We provide more details on data collection and statistics in the following context.

\begin{table}[ht]
\small
    \centering
    \begin{tabular}{ll|ll}
    \toprule
        Type & Number & Type & Number \\
    \midrule
         Long-sleeve coat & 157 &  Long-sleeve dress & 18\\
         Short-sleeve coat & 98 & Short-sleeve dress & 34\\ 
         None-sleeve coat & 35 & None-sleeve dress & 32 \\
         Long trousers & 29 & Long skirt & 104\\
         Short trousers & 44 & Short skirt & 48\\
    \bottomrule
    \end{tabular}
    \caption{Statistics of the each clothing categories of \datasetName{}.}
    \label{tab:dataset}
\end{table}

\begin{figure}[ht]
\centering
\includegraphics[width=1.0\linewidth]{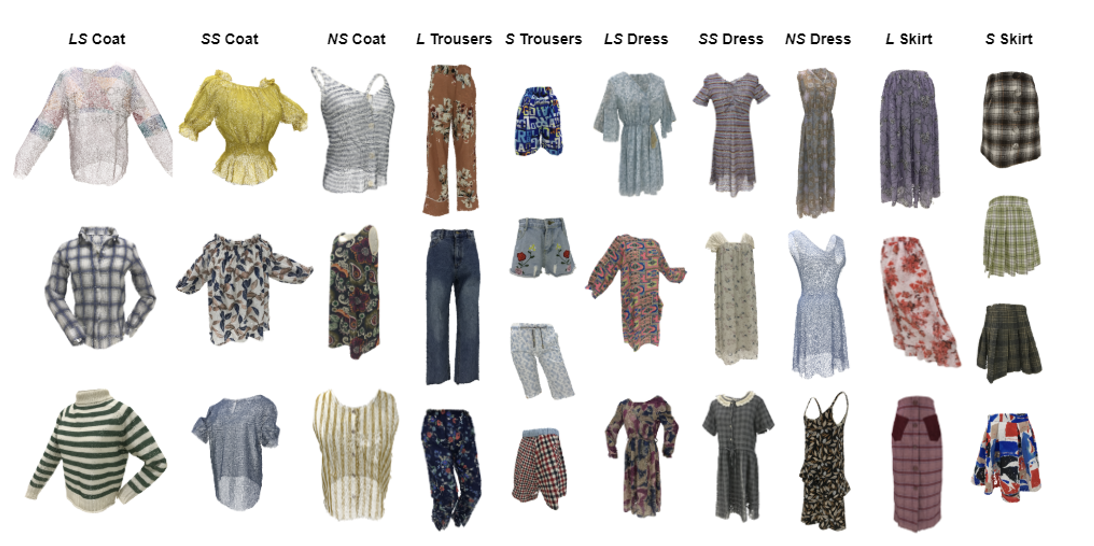}
\caption{Example garment models of \datasetName{}. }
\label{fig:dataset}
\end{figure}

\paragraph{Cloth Capture.}
To model the large variety of real-world clothing, we collect a large number of garments, consisting of 563 diverse  items that covers 10 clothing categories: long/short/none-sleeve coat, long/short/none-sleeve dress, long/short trousers and long/short skirt. The detailed numbers for each category are shown in Table~\ref{tab:dataset}.
We adopt the image-based geometry reconstruction software~\cite{Mentashape} to generate high-resolution garment
reconstructions from multi-view images in the form of dense point cloud.
In particular, the input images are captured in a multi-view studio with of 50 RGB cameras and controlled lighting.  
To enhance the expressiveness of the dataset, each garment item is randomly posed on a dummy model or real human to generate a large variety of real deformations caused by body motion.
With the augmentation of poses, 2078 3D garment models in total are reconstructed from our pipeline.



\begin{figure}
\begin{floatrow}
\ffigbox{%
  \includegraphics[scale=0.2]{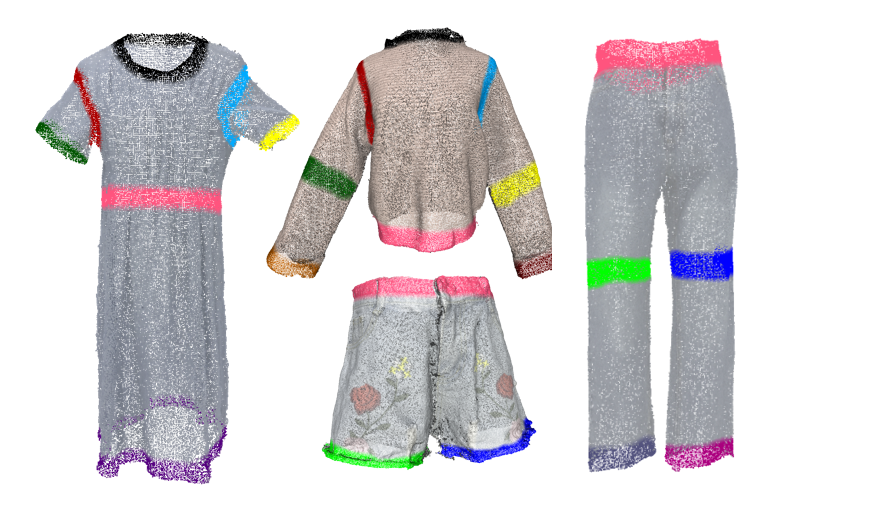}%
}{%
  \caption{Visualization of feature line annotations. Different feature lines are highlighted using different colors.}%
  \label{fig:featureLine}
}
\capbtabbox{%
\small{
  \begin{tabular}{cc} \hline
  Cloth Category & Feature line Positions \\ \hline
  long-sleeve coat & ne, wa, sh, el, wr\\ 
  short-sleeve coat & ne, wa, sh, el\\
  none-sleeve coat &  ne, wa, sh\\
  long-sleeve dress &  ne, wa, sh, el, wr, he\\ 
  short-sleeve dress & ne, wa, sh, el, he\\
  none-sleeve dress &  ne, wa, sh, he\\ 
  long/short trousers &  wa, kn, an/ wa, kn\\ 
  long/short skirt & wa, he/ wa, he \\ 
  \hline
  \end{tabular}
}}{%
  \caption{Feature line positions for each cloth category. The meanings for the abbreviations are: 'ne'-neck, 'wa'-waist, 'sh'-shoulder, 'el'-elbow, 'wr'-wrist, 'kn'-knee, 'an'-ankle, 'he'-'hemline'. }%
  \label{tab:featureLine}
}
\end{floatrow}
\end{figure}

\paragraph{Annotations.}
To facilitate future research on 3D garment reasoning, apart from the calibrated multi-view images, we provide additional annotations for \datasetName{}.
In particular, we introduce \textit{feature line} annotation which is specially tailored for 3D garments. Akin to facial landmarks, the feature lines denote the most prominent features of interest, e.g. the open boundaries, the neckline, cuff, waist, etc, that could provide strong priors for a faithful reconstruction of 3D garment.
The details of feature line annotations are provided in Table~\ref{tab:featureLine} and visualized in Figure~\ref{fig:featureLine}. We will show in method section that those feature line labels can supervise the learning of 3D key lines prediction, which provide explicit constraints for mesh generation. 

Furthermore, each reconstructed model is labeled with 3D pose represented by SMPL~\cite{loper2015smpl} coefficients. The pose is obtained by fitting the SMPL model to the reconstructed dense point cloud.  Due to the highly coupled nature between human body and clothing, we believe the labeled 3D pose could be beneficial to infer the global shape and pose-dependent deformations of the garment model.



\begin{table}[ht]
\centering
\scriptsize
    \begin{tabular}{lcccc}
    \toprule
          & Wang et al.~\cite{garmentdesign_Wang_SA18} & GarNet~\cite{gundogdu2019garnet} & MGN~\cite{bhatnagar2019multi} & \textbf{\datasetName{}} \\
    \midrule
         $\#$ Models &  2000 & 600 & 712 & 2078 \\ 
         $\#$ Categories & 3 & 3 & 5 & 10 \\ 
         Real/Synthetic &  synthetic & synthetic & real & real \\ 
         Method  & simulation & simulation & scanning & multi-view stereo \\ 
    \midrule
        Annotations & input 2D sketch & 3D body pose & \begin{tabular}{@{}c@{}} vertex color \\ 3D body pose \end{tabular} & \begin{tabular}{@{}c@{}} multi-view real images \\ 3D feature lines \\ 3D body pose \end{tabular} \\
    \bottomrule
    \end{tabular}
    \caption{Comparisons with other 3D garment datasets.}
    \label{tab:data_comp}
\end{table}

\paragraph{Data Statistics.}
To the best of our knowledge, among existing works, there are only three publicly available datasets specialized for 3D garments: Wang et. al~\cite{garmentdesign_Wang_SA18}, GarNet~\cite{gundogdu2019garnet} and MGN~\cite{bhatnagar2019multi}.
In Table~\ref{tab:data_comp}, we provide detailed comparisons with these datasets in terms of the number of models, categories, data modality, production method and data annotations. 
Scale-wise, \datasetName{} and Wang et al.~\cite{garmentdesign_Wang_SA18} are one order larger than the other counterparts. However, our dataset covers much more garment categories compared to Wang et al.~\cite{garmentdesign_Wang_SA18}.
Apart from our dataset, only MGN collects models reconstructed from real garments while the other two are fully synthetic.
Regarding data annotations, \datasetName{} provides the richest data labels. In particular, multi-view real images are only available in our dataset. In addition, we present a new form of garment annotation, the 3D feature lines, which could offer important landmark information for a variety of 3D garment reasoning tasks including garment reconstruction, segmentation, retrieval, etc.

	\section{A Baseline Approach for Single-view Reconstruction}
\label{sec:method}

To demonstrate the usefulness of \datasetName{}, we propose a novel baseline approach for single-view garment reconstruction. Specifically, taking a single image $I$ of a garment as input, we aim to reconstruct its 3D shape represented as a triangular mesh. 
Although recent advances in 3D deep learning techniques have achieved promising progress in single-view reconstruction on general objects, we found all existing approaches have difficulty scaling to cloth reconstruction.
The main reasons are threefolds: 
(1) \emph{Non-closed surfaces.} 
Unlike the general objects in ShapeNet~\cite{chang2015shapenet}, the garment shape  typically appears as a thin layer with open boundary.  
While implicit representation~\cite{mescheder2019occupancy,park2019deepsdf} can only model closed surface, voxel based approach~\cite{choy20163d} is not suited for recovering shell-like structure like the garment surface.
(2) \emph{Complex shape topologies.} 
As all existing mesh-based approaches~\cite{groueix2018,wang2018pixel2mesh,pan2019deep} rely on deforming a fixed template, they fail to handle the highly diversified topologies introduced by different clothing categories.
(3) \emph{Complicated geometric details.} 
While general man-made objects typically consist of smooth surfaces, the clothing dynamics often introduces intricate high-frequency surface deformations that are challenging to capture.

\begin{figure*}[t]
\centering
\includegraphics[width=1\linewidth]{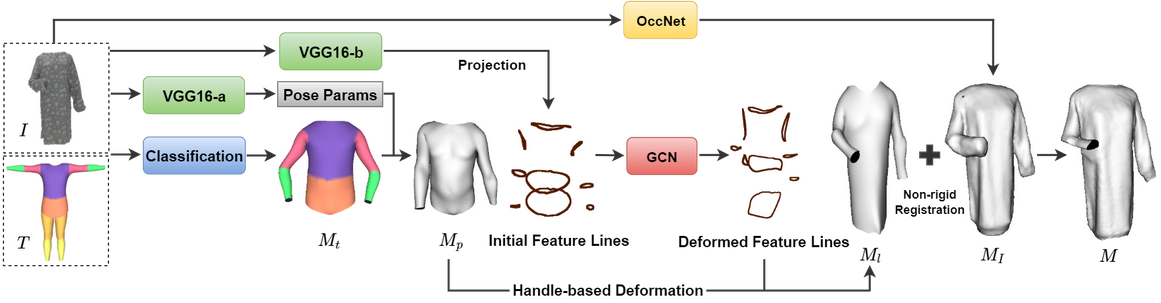}
\caption{The pipeline of our proposed approach. }
\label{fig:pipeline}
\end{figure*}



\paragraph{Overview.} To address the above issues, we propose to employ a hybrid representation that leverages the merits of each embedding. 
In particular, we harness both the capability of implicit surface of modeling fine geometric details and the flexibility of mesh representation of handling open surfaces.
Our method starts with generating a template mesh $M_t$ which can automatically adapt its topology to fit the target clothing category in the input image. It is then deformed to $M_p$ by fitting the estimated 3D pose. By treating the feature lines as a graph, we then apply image-guided graph convolutional network (GCN) to capture the 3D feature lines, which later trigger handle-based deformation and generates mesh $M_l$. 
To exploit the power of implicit representation, we first employ OccNet ~\cite{mescheder2019occupancy} to generate a mesh model $M_I$ and then adaptively register $M_l$ to $M_I$ by incorporating the learned fine surface details from $M_I$ while discarding its outliers and noises caused by enforcement of close surface.
The proposed pipeline is illustrated in Figure~\ref{fig:pipeline}. 



\subsection{Template Mesh Generation}
\label{sec:template}


\paragraph{Adaptable template.}
We propose \emph{adaptable template}, a new representation that is scalable to different cloth topologies, enabling the generation of all types of cloth available in the dataset using a single network.
The adaptable template is built on the  SMPL~\cite{loper2015smpl} model by removing the head, hands and feet regions.  
As seen in Figure~\ref{fig:pipeline}, it is then segmented into 6 semantic regions: torso, waist, and upper/lower limbs/legs. 
During training, the entire adaptable template is fed into the pipeline. However, different semantic regions are activated according to the estimated cloth topology.
We denote the template mesh as $M_t = (V, E, B)$, where $V=\{v_i\}$ and $E$ are the set of vertices and edges respectively, and $B = \{b_i\}$ is a per-vertex binary activation mask. 
$v_i$ will only be activated if $b_i=1$; otherwise $v_i$ will be detached during the training and removed in the output. 
The activation mask is determined by the estimated cloth category, where regions of vertices are labeled as a whole.
For instance, to model a short-sleeve dress, vertices belonging to the regions of lower limbs and legs are deactivated. 
Note that in order to adapt the waist region to large deformations for modeling long dresses, we densify its triangulation accordingly using mesh subdivisions.

\paragraph{Cloth classification.} 
We build a cloth classification network based on a pre-trained VGGNet. 
The classification network is trained using both real and synthetic images. 
The synthetic images are used in order to provide augmented lighting conditions to the training images.
In particular, we render each garment model under different global illuminations in 5 random views.
We generate around 10,000 synthetic images, 90\% of which is used for training while the rest is reserved for testing. 
Our classification network can achieve an accuracy of 99.3\%, leading to an appropriate template at both train and test time.

\subsection{Learning Surface Reconstruction}
\label{sec:recon}


To achieve a balanced trade-off between mesh smoothness and accuracy of reconstruction, we propose a multi-stage pipeline to progressively deforming $M_t$ to fit the target shape. 

\subsubsection{Feature line-guided Mesh Generation.}
\label{sec:feature_fit}

It is well understood that, the feature lines, such as necklines, hemlines, etc, play a key role in casting the shape contours of the 3D clothing.
Therefore, we propose to first infer the 3D feature lines and then deform $M_t$ by treating the feature lines as deformation handles.


\paragraph{Pose Estimation.}
\label{sec:posefit}

Due to the large degrees of freedom of 3D lines, directly regressing their positions is highly challenging. To reduce the searching space and thus make the problem tractable, we firstly estimate the body pose and deform $M_t$ to obtain a new mesh $M_p$ which provides an initialization $\{l^p_i\}$ of 3D feature lines.  
Here, the pose of 3D garment is represented as the SMPL pose parameters $\theta$ \cite{loper2015smpl}, which are regressed by a pose estimation network. 

\paragraph{GCN-based Feature line regression.}

We represent the feature lines $\{l^p_i\}$ as polygons during pose estimation.
This enables us to treat it as a graph and further employ an image-guided GCN to regress the vertex-wise displacements. We employ another VGG module to extract the features from the input image and leverage a similar learning strategy with Pixel2Mesh~\cite{wang2018pixel2mesh} to infer deformation of feature lines.
Note that before the regression step, we first determine the activated subset of feature lines according to the estimated cloth category and only feed the activated ones into the network.


\paragraph{Handle-based deformation.} 
We denote the output feature lines of the above steps as $\{l^o_i\}$. 
$M_l$ is obtained by deforming $M_p$ so that its feature lines $\{l^p_i\}$ fit our prediction $\{l^o_i\}$. 
We use the handle-based Laplcacian deformation~\cite{sorkine2004laplacian} by setting the alignment between $\{l^p_i\}$ and $\{l^o_i\}$ as hard constrains while optimizing the displacements of the remaining vertices to achieve smooth and visually pleasing deformations. 
Note that the explicit handle-based deformation can quickly lead to a result that is close to the target surface, which alleviates the difficulty of regressing of a large number of vertices.

\subsubsection{Surface Refinement by Fitting Implicit Reconstruction.}
\label{sec:surf_refine}
After obtaining $M_l$, a straightforward way to obtain refined surface details is to apply Pixel2Mesh ~\cite{wang2018pixel2mesh} by directly taking $M_l$ as input. However, as illustrated in Fig.~\ref{fig:qualiResults}, this method fails probably due to the inherent difficulty of learning the high-frequency details while preserving surface smoothness. 
In contrast, our empirical results indicate that the implicit surface based methods, such as OccNet~\cite{mescheder2019occupancy}, can faithfully recover the details but only generate closed surface.  
We therefore directly perform an adaptive non-rigid registration from $M_l$ to the output of OccNet for transferring the learned surface details. 

\paragraph{Learning implicit surface.} We directly employ OccNet~\cite{mescheder2019occupancy} for learning the implicit surface. Specifically, the input image is first encoded into a latent vector using ResNet-18. For each 3D point in the space, a MLP layer consumes its coordinate and the latent code and predict if the point is inside or outside the surface. 
Note that we convert all the data into closed meshes using Poisson reconstruction method. With the trained network, we first generate an implicit field and then extract the reconstructed surface $M_I$ using marching cube algorithm~\cite{lorensen1987marching}.

\paragraph{Detail transfer with adaptive registration.} 
Though OccNet can synthesize high-quality geometric details, it may also introduce outliers due to its enforcement of generating closed surface. We therefore propose an adaptive registration to only transfer correct high-frequency details by imposing two additional constraints to the conventional non-rigid ICP algorithm:
(1) the two points of a valid correspondence should have consistent normal direction (i.e., the angle of the two normal directions should be smaller than a threshold which is set as $60^\circ$). (2) the bi-directional Chamfer distance between the corresponded points should be less than a preset threshold $\sigma$ ($\sigma$ is set as 0.01). The adaptive registrations helps to remove erroneous correspondences and produces our final output $M_r$.

\subsection{Training}
\label{sec:training}

There are four sub-networks need to be trained: cloth classification, pose estimation, GCN-based feature line fitting and the implicit reconstruction. Each of the sub-networks is trained independently. 
In the following subsections, we will provide the details on training data preparation and loss functions. 

\subsubsection{Training Data Generation}

\paragraph{Pose estimation.} 
We obtain the 3D pose of the garment model by fitting the SMPL model to the reconstructed dense point cloud. 
The data processing procedures are as follows:
1) for each annotated feature line, we calculate its center point as the its corresponding skeleton joint; 2) we use the joints in the torso region to align all the point clouds to ensure a consistent orientation and scale.
3) lastly, we compute the SMPL pose parameters for each model by fitting the joints and point cloud. 
The obtained pose parameters will be used for supervising the pose estimation module in Section~\ref{sec:posefit}.

\paragraph{Image rendering.} 
We augment the input with synthetic images.
In particular, for each model, we generate rendered images by randomly sampling 3 viewpoints and 3 different lighting environments, obtaining 9 images in total. 
Note that we only sample viewpoints from the front viewing angles as we only focus on front-view reconstruction in this work. 
However, our approach can easily scale to side or back view prediction by providing the approach according training images. 


\subsubsection{Loss functions}
The training of cloth classification, pose estimation and implicit reconstruction exactly follows the mainstream protocols. Hence, due to the page limit, we only focus on the part of feature line regression here while leaving other details in the appendix.  

\paragraph{Feature line regression.} Our training goal is to minimize the average distance between the vertices on the obtained feature lines and the ground-truth annotations. Therefore, our loss function is a weighted sum of a distance metric (we use Chamfer distance here) and an edge length regularization loss \cite{wang2018pixel2mesh}, which helps to smooth the deformed feature lines.
More details can be found in the appendix.

\begin{figure*}[h]
\centering
\includegraphics[width=0.95 \linewidth]{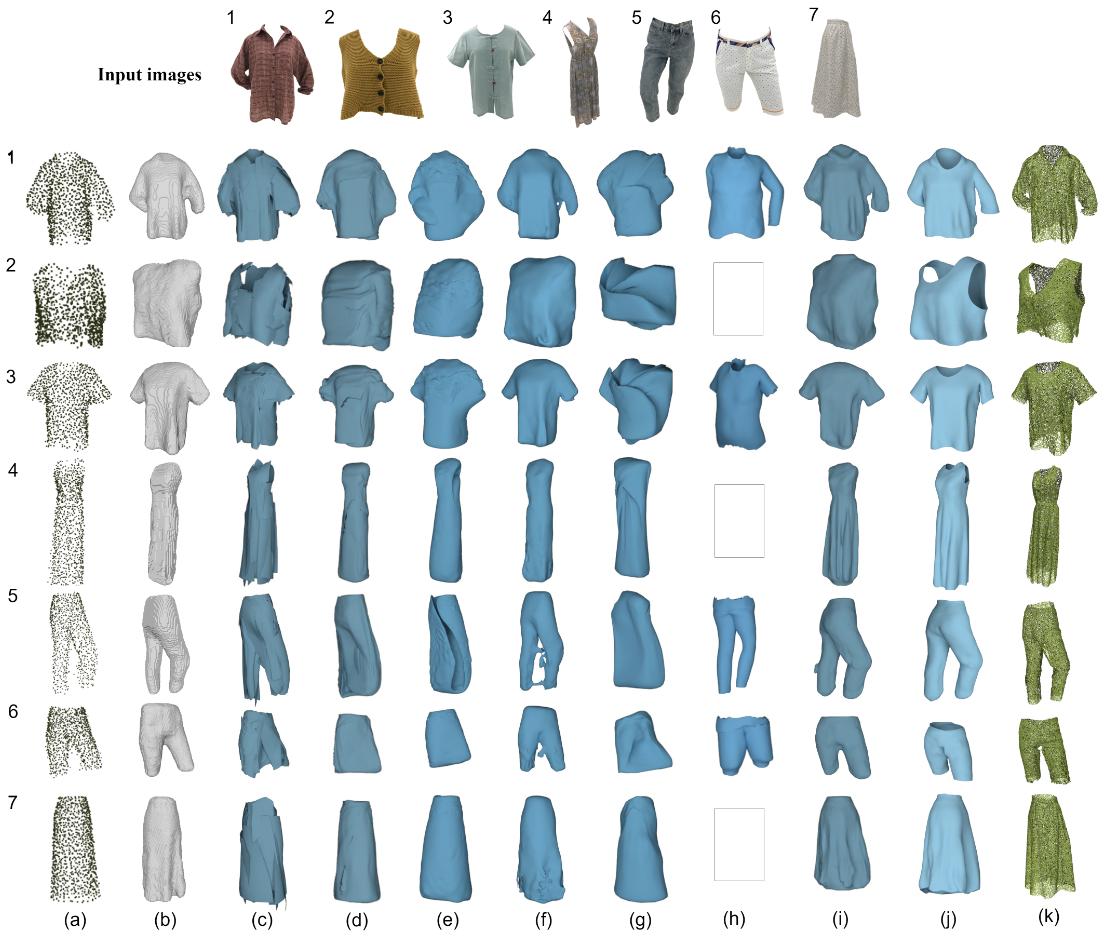}
\caption{Experiment results against other methods. Given an image, results are followed with (a) PSG (Point Set Generation) \cite{Fan_2017_CVPR}; (b) 3D-R2N2 \cite{choy20163d};  (c) AtlasNet\cite{groueix2018} with 25 square patches; (d) AtlasNet\cite{groueix2018} whith a sphere template; (e) Pixel2Mesh \cite{wang2018pixel2mesh}; (f) MVD \cite{lun20173d} (multi-view depth generation); (g) TMN \cite{pan2019deep} (topology modification network); (h) MGN (Multi-Garment Network) \cite{bhatnagar2019multi}; (i) OccNet \cite{mescheder2019occupancy}; (j) Ours; (k) The groundtruth point clouds. The input images on the top. The null means the method fails to generate a result. 
}
\label{fig:qualiResults}
\end{figure*}

\section{Experimental Results}
\label{sec:result}
  
\paragraph{Implementation details.}
The whole pipeline proposed is implemented using PyTorch. The initialized learning rate is set to 5e-5 and with the batch size of 8. It takes about 30 hours to train the whole network using Adam optimization for 50 epochs using a NVIDIA TITAN XP graphics card.

\subsection{Benchmarking on Single-view Reconstruction}


\paragraph{Methods.}
We compare our method against six state-of-the-art single-view reconstruction approaches that use different 3D representations: 3D-R2N2 \cite{choy20163d}, PSG(Point Set Generation) \cite{Fan_2017_CVPR}, MVD (generating multi-view depth maps) \cite{lun20173d}, Pixel2Mesh \cite{wang2018pixel2mesh}, AtlasNet \cite{groueix2018}, MGN \cite{bhatnagar2019multi} and OccNet~\cite{mescheder2019occupancy}. 
For AtlasNet, we have experimented it using both sphere template and patch template, which are denoted as ``Atlas-Sphere'' and ``Atlas-Patch'' (25 square patches are used). 
To ensure fairness, we train all the algorithms, except MGN, on our dataset. 
In particular, training MGN requires ground-truth parameters for their category-specific cloth template, which is not applicable in our dataset.
It is worth mentioning that, the most recent algorithm MGN can only handle 5 cloth categories and fails to produce reasonable results for out-of-scope classes, e.g., dress, as demonstrated in Fig.~\ref{fig:qualiResults}. 
To obtain the results of MGN, we manually prepared input data to fulfill the requirements of its released model, that is trained on digital wardrobe \cite{bhatnagar2019multi}.   


\paragraph{Quantitative results.}
Since the approaches leverage different 3D representations, we convert their output into point cloud to ensure fair comparison.
We then compute the Chamfer distance (CD) and Earth Mover's distance (EMD) between the outputs and the ground-truth point clouds for quantitative measurements.
Table~\ref{table:quantitative} shows the performance of different methods on our testing dataset. 
Our approach achieves the highest reconstruction accuracy compared to the other approaches. 

\begin{table}
	\setlength{\tabcolsep}{0.25cm}
	\centering
	\begin{tabular}{l | c | c }
		\toprule
		Method           & CD($\times 10^{-3}$) & EMD ($\times 10^{2}$) \\
		\midrule
		3D-R2N2 ($128^3$) \cite{choy20163d}          & 1.264 & 3.609 \\
		MVD \cite{lun20173d}      & 1.047 &  4.058   \\
		PSG \cite{Fan_2017_CVPR}        & 1.065 &  4.675  \\
        Pixel2Mesh \cite{wang2018pixel2mesh}      & 0.782 & 9.078\\
		AtlasNet(sphere)  \cite{groueix2018}   & 0.855 & 6.193  \\
		AtlasNet(patch) \cite{groueix2018}    & 0.908 & 9.428\\
		TMN \cite{pan2019deep} & 0.865 & 8.580\\
		OccNet ($256^3$) \cite{mescheder2019occupancy} & 0.960 & 3.431 \\
		Ours & \textbf{0.679} & \textbf{2.942} \\
		\bottomrule
	\end{tabular}
	\caption{The prediction errors of different methods evaluated on our testing data.}
	\label{table:quantitative}
\end{table}
 
\paragraph{Qualitative results.}
I n Figure~\ref{fig:qualiResults}, we also provide qualitative comparisons by randomly selecting some samples from different garment categories in arbitrary poses.
Compared to the other methods, our approach provides more accurate reconstructions that are closer to ground truths. The reasons are: 1) 3D representations like point set \cite{Fan_2017_CVPR}, voxel \cite{choy20163d} or multi-view depth maps \cite{lun20173d} are not suitable for generating a clean mesh. 2) Although template-based methods \cite{groueix2018,wang2018pixel2mesh,pan2019deep} are designed for mesh generation, it is hard to use a fixed template for fitting diverse shape complexity of clothing. 3) As shown in the results, method based on implicit function~\cite{mescheder2019occupancy} is able to synthesis rich details. However, it can only generate closed shapes, making it difficult to handle garment reconstruction, which typically consists of multiple open boundaries.
By explicitly combining the merits of template-based methods and implicit ones, the proposed approach can not only capture the global shape but also generate faithful geometric details.         


\begin{figure}[ht]
\centering
\includegraphics[width=0.95\linewidth]{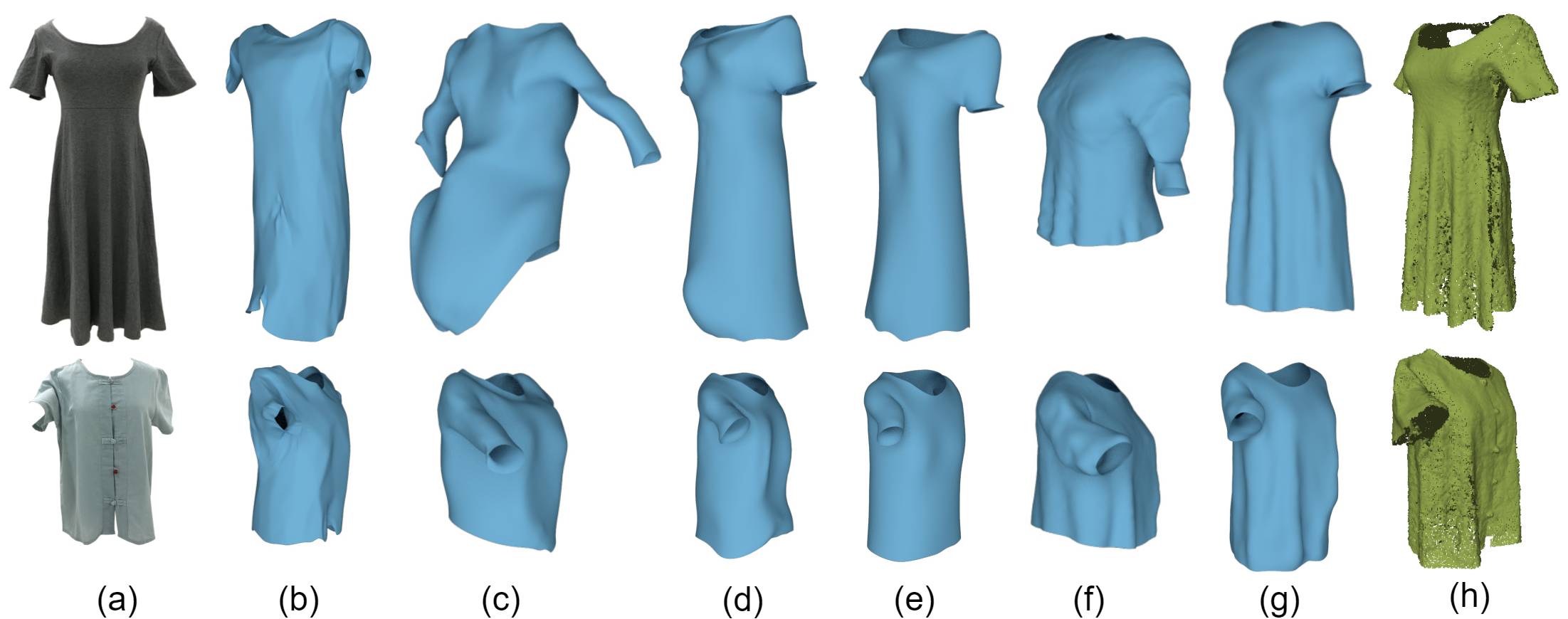}
\caption{Results of ablation studies. (a) input images; (b) results of $M_t$+GCN; (c) results of $M_p$+GCN; (d) results of $M_l$+GCN. (e) results of our approach without surface refinement, i.e., $M_l$. (f) $M_t$+Regis. (g) results of our full approach. (h) groundtruth point clouds. }
\label{fig:ablation}
\end{figure}

\subsection{Ablation Analysis}
We further validate the effectiveness of each algorithmic component by selectively applying them in different settings: 1) Directly applying GCN on the generated template mesh $M_t$ for fitting the target shape, termed as $M_t$+GCN; 2) Applying GCN on $M_p$ (obtained by deforming $M_t$ with estimated SMPL pose) to fit the target shape, termed as $M_p$+GCN; 3) Applying GCN on the resulted mesh after feature line-guided deformation, i.e. $M_l$. This is termed as $M_l$+GCN; 4) Directly performing registration from $M_t$ to $M_I$ for details transferring, which is termed as $M_t$+Regis. Figure~\ref{fig:ablation} shows the qualitative comparisons between these settings and the proposed one. As seen, the baseline approach produce the best results.

As observed from the experiments, it is difficult for GCN to learn geometric details. There are two possible reasons: 1) It is inherently difficult to synthesize high-frequency signals while preserving the surface smoothness; 2) GCN structure might be not suitable for a fine-grained geometric learning task as graph is a sparse and crude approximation of a surface. We also found that the featurelines are much easier to learn and an explicit handle-based deformation works surprisingly well. The deeper study in this regard is left as one of our further works.

	\section{Conclusions and Discussions}
\label{sec:conclusion}

We have proposed a new dataset and benchmark, called \datasetName{}, for image-based garment reconstruction.
To the best of our knowledge, \datasetName{} is the largest collection by far of 3D garment models reconstructed from real clothing images.
In particular, it consists of over 2000 highly diversified garment models covering 10 clothing categories and 563 distinct garment items. In addition, each model of \datasetName{} is richly labeled with 3D body pose, 3D feature lines and multi-view real images. 
We also presented a baseline approach for single-view reconstruction to validate the usefulness of the proposed dataset. 
It uses a novel representation, called adaptable template, to learn a variety of clothing types in a single network. 
We have performed extensive benchmarking on our dataset using a variety of recent methods. We found that single-view garment reconstruction is an extremely challenging problem with ample opportunity for improved methods. We hope \datasetName{} and our baseline approach will bring some insight to inspire future research in this field.

    \section*{Acknowledgment}
\noindent The work was supported in part by grants No. 2018YFB1800800, No. 2018B030
338001, No. 2017ZT0 7X152, No. ZDSYS201707251409055 and in part by National Natural Science Foundation of China (Grant No.: 61902334 and 61629101). The authors also would like to thank Yuan Yu for her early efforts on dataset construction.

	


	\appendix
	\clearpage
	\section*{Appendix}
	In this appendix, we provide more details, more results and more analysis in the following aspects: 1) more implementation details on the training settings of cloth classification, pose estimation, feature line regression and the design of the adaptable template; 2) more experiments demonstrating the effectiveness of the adaptable template; 3) more results reconstructed from the in-the-wild images using by our baseline method.
\setlength{\textfloatsep}{4pt}
\section{Implementation Details}


In the main paper, we have briefly introduced the losses adopted to train each stage for our proposed approach. In this section, we will describe the detailed training setting of cloth classification, pose estimation, implicit reconstruction along with the loss functions and the hyper parameters used in the feature line regression module.

\subsection{Cloth Classification}
We build a cloth classification network to infer the type of clothes presented in input images, which consists of a VGG16 feature extractor and a fully connected layer. The network is trained using a standard cross-entropy loss. 
To enhance the generalization performance of our model, the cloth classification module is trained on a hybrid source combining synthetic images rendered using the point clouds of our dataset, the real images selected from DeepFashion~\cite{liuLQWTcvpr16DeepFashion} and DeepFashion2~\cite{ge2019deepfashion2}, and the real multi-view images from our \datasetName{}. 
We use 90\% of the data for training while leaving the other 10\% for testing. Random resized crop as well as random rotation are employed to augment the data during training. Furthermore, we balanced the number of images used for training by randomly selecting clothes from each category, so that for each cloth category, the number of images involved in the training is nearly the same.

\subsection{Pose Estimation}
As our adaptable template is built upon SMPL~\cite{loper2015smpl}, we employ a subset of SMPL parameters to denote the pose of the clothes while setting the parameters irrelevant to the deformation of the clothes, such as global rotation, local rotation parameters for ankle, wrist, neck, etc., to zero. 
To estimate the partial pose parameters from the image, we build a pose estimation module consisting of a VGG16 feature extractor and a fully connected layer.

We train the pose estimation network by minimizing the loss function $\mathcal{L}_{pose}$\ which consists of a cloth pose parameter loss $\mathcal{L}_{param}$\ and a pose regularization loss $\mathcal{L}_{reg}$\ :

\begin{center}
    $\mathcal{L}_{pose}$\ = $\mathcal{L}_{param}$\ + $\lambda_{reg}$\ $\mathcal{L}_{reg}$,\
\end{center}

The pose parameter loss $\mathcal{L}_{param}$\ is calculated as the MSE between the ground truth and the predicted pose parameters. 
We further introduce a regularization loss $\mathcal{L}_{reg}$\ that is the squared sum of the pose parameters, aiming to eliminate unexpected rotations. 
During training, $\lambda_{reg}$\ is set to 1e-5.

\subsection{Implicit Surface Reconstruction}

We adopt OccNet \cite{mescheder2019occupancy} conditioned on image features for implicit reconstruction.
The feature extraction module consists of a pre-trained ResNet-18 which is fine-tuned using the real and synthetic cloth images. 
A fully connected layer is used to predict whether the input 3D point is inside the surface or not given its coordinate and a latent code of the image. We exactly follow the settings of the original OccNet for training.

\subsection{Feature Line Regression} 

As mentioned in the main paper, we proposed a novel feature line loss $\mathcal{L}_{line}$\ as well as the edge length regularization loss $\mathcal{L}_{edge}$\ to guide the feature line synthesis while reducing the zigzag artifacts: 

\begin{center}
    $\mathcal{L}_{fitting}$\ = $\mathcal{L}_{line}$\ + $\lambda_{edge}$\ $\mathcal{L}_{edge}$.\
\end{center}

\noindent 
The feature line loss $\mathcal{L}_{line}$\ is calculated using Chamfer distance, which measures the average squared distance between the vertices on the predicted feature lines and the corresponding feature lines annotated on the ground-truth point clouds. 
The edge length regularization term is calculated as the average squared length of the edges on the selected feature lines.
It is worth mentioning that the feature lines on the adaptable template are a series of closed polygonal curves while the "feature lines" of the ground truth point clouds are a subset of the points that lie around the intended landmark regions (see Fig.~\ref{fig:moreexperiments} right). During training,  $\lambda_{edge}$\ is set to 0.2 according to our experiment to strike a balance between reconstruction accuracy and smoothness of the generated feature lines.

\begin{figure}[h]
\centering
\includegraphics[width=0.80\linewidth]{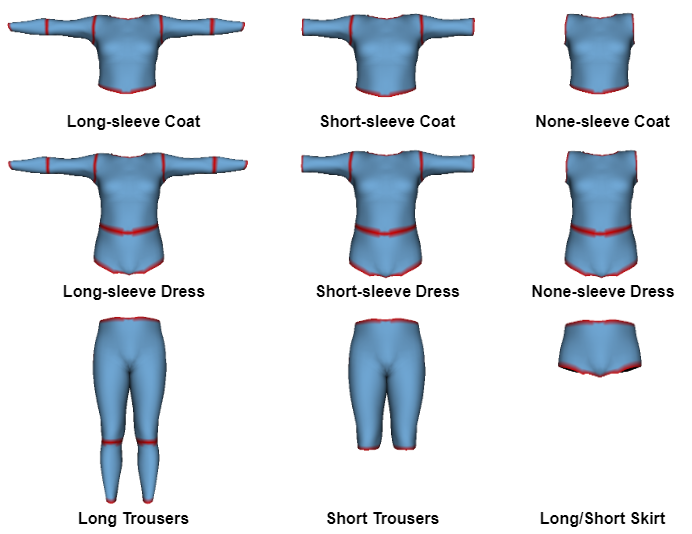}
\caption{Activated semantic regions and feature lines of adaptable template for each type of clothes. The feature lines are colored in red. Long and short skirts share the same activated patterns.}
\label{fig:templatefigure}
\end{figure}

\subsection{Adaptable Template} 

In this section, we provide more details regarding the generation of the adaptable template. Fig.~\ref{fig:templatefigure} shows the activated semantic regions of our adaptable template when dealing with different types of garments. The curves colored in red are the activated feature lines for each scenario.



\section{Evaluation of Adaptable Template}


In this section, we further evaluate the effectiveness of the proposed adaptable template by comparing it with an alternative method that uses type-specific template.
In particular, the comparing method leverages different models for reconstructing different types of clothing -- each model is trained using one type-specific cloth template and its corresponding subset of the 3D data and training images.
In contrast, as our adaptable template can handle all available clothing categories in \datasetName{}, our approach is trained on the entire training set.
Note that, to ensure fair comparison, the training of these two approaches share the identical settings, network structure, and losses except the above-mentioned differences in the training data.

\begin{figure}[h]
\centering
\includegraphics[width=0.75\linewidth]{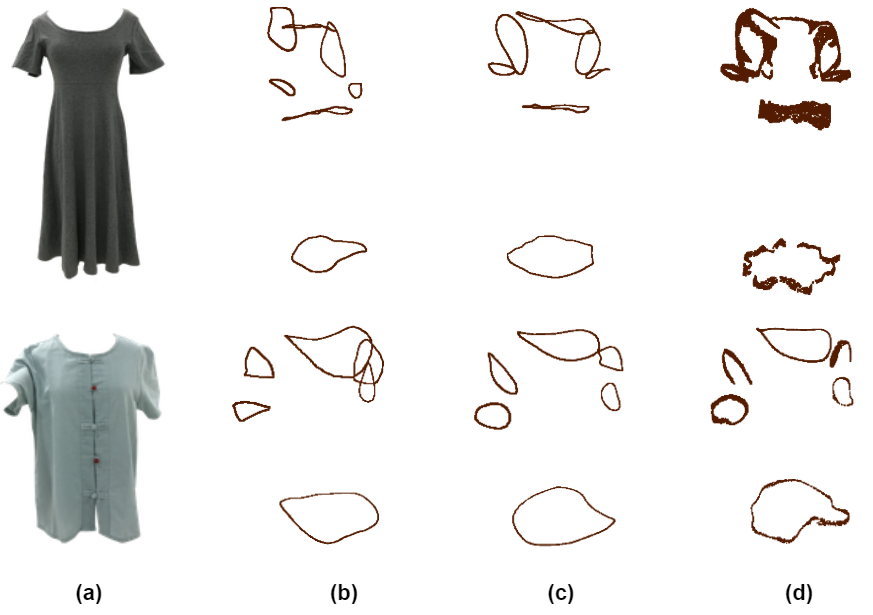}
\caption{
Comparisons with category-specific models in terms of feature line prediction.
Given the input images (a) on the left, we show (b) feature lines generated by method using type-specific template and trained with type-specific data; (c) feature lines generated using our approach with adaptable template and trained with full training data covering all clothing categories; (d) ground truth. Note that the ground-truth feature lines are labeled on reconstructed point clouds which are also point clouds as shown here.}
\label{fig:moreexperiments}
\end{figure}

We compare the performance of two approaches in terms of feature line prediction on novel images as shown in Figure~\ref{fig:moreexperiments}. 
Our approach can generate much more plausible feature lines compared to that of the approach which relies on category-specific templates.
In addition, our predicted features lines are very close to the ground truth despite the different data modalities. 
The stronger generalization performance indicates the advantages of our proposed adaptable template which is able to harness all kinds of training data at a single network to prevent overfitting.



\begin{figure}[h]
\centering
\includegraphics[width=0.85\linewidth]{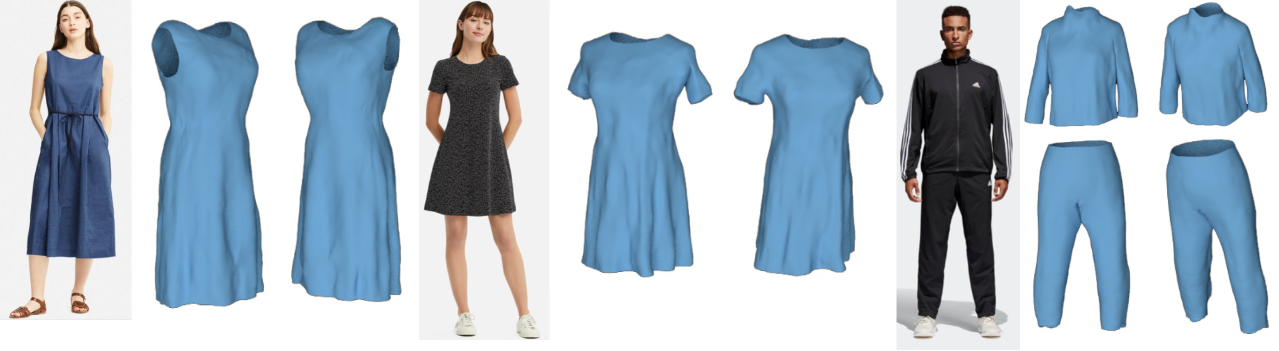}
\caption{Reconstruction from in-the-wild images using our approach. For each result, we visualize the obtained reconstructed meshes in two different views.}
\label{fig:realimageresults}
\end{figure}
\vspace{-1em}
\section{More Results on In-the-wild Images}
In this section, we show more results generated by our baseline approach trained on our \datasetName{} dataset in Figure~\ref{fig:realimageresults}.
Note that all the images adopted for testing are in-the-wild images from the Internet, which are not seen during training.
As seen from the results, our approach can well capture the clothing topologies given a variety of input styles while faithfully recovering the geometric details.

	
	\bibliographystyle{splncs04}
	\bibliography{paper}
	
	\ifthenelse{\equal{\final}{0}}
	{
	}

	{}
\end{document}